\pdfoutput=1

\documentclass[11pt]{article}

\usepackage{arxiv}
\usepackage[numbers]{natbib}
\usepackage[table]{xcolor}
\usepackage{times}
\usepackage{latexsym}
\usepackage[T1]{fontenc}
\usepackage[utf8]{inputenc}

\usepackage{microtype}
\IfFileExists{inconsolata.sty}{\usepackage{inconsolata}}{}%
\usepackage{graphicx}
\usepackage{float}              %
\usepackage{booktabs}           %
\usepackage{multirow}
\usepackage{enumitem}
\usepackage{alltt}
\usepackage{pgfplots}           %
\pgfplotsset{compat=1.18}

\newcommand{\mask}{\texttt{[MASK]}}
\newcommand{\pad}{\texttt{[PAD]}}
\newcommand{\ours}{\textsc{SDDC}}

\newcommand{\hlg}[1]{{\setlength{\fboxsep}{1pt}\colorbox{green!20}{\scriptsize #1}}}   %
\newcommand{\hlly}[1]{{\setlength{\fboxsep}{1pt}\colorbox{yellow!35}{\scriptsize #1}}} %
\newcommand{\hlr}[1]{{\setlength{\fboxsep}{1pt}\colorbox{red!15}{\scriptsize #1}}}     %
\newcommand{\lat}[1]{\textcolor{green!45!black}{#1}}                                   %
\newcommand{\spd}[1]{{\scriptsize(#1$\times$)}}                                        %

\usepackage{amsmath,amsfonts,bm}

\def\eqref#1{equation~\ref{#1}}

\def\1{\bm{1}}

\def\vc{{\bm{c}}}

\def\vw{{\bm{w}}}
\def\vx{{\bm{x}}}
\def\vy{{\bm{y}}}
\def\vz{{\bm{z}}}

\DeclareMathAlphabet{\mathsfit}{\encodingdefault}{\sfdefault}{m}{sl}
\SetMathAlphabet{\mathsfit}{bold}{\encodingdefault}{\sfdefault}{bx}{n}

\def\gV{{\mathcal{V}}}

\newcommand{\E}{\mathbb{E}}

\newcommand{\KL}{D_{\mathrm{KL}}}

\usepackage{amssymb}
\usepackage{amsthm}
\newtheorem{proposition}{Proposition}
\usepackage{algorithm}
\usepackage[noend]{algpseudocode}
\renewcommand{\eqref}[1]{\textup{(\ref{#1})}}

\newcommand{\narrowtablewidth}{0.5\textwidth}
\newcommand{\arxivversion}{}
\newcommand{\teaseraxiswidth}{4.4cm}
\newcommand{\teaseraxisheight}{3.0cm}

\usepackage{hyperref}

\allowdisplaybreaks

\title{Self-Decomposing Diffusion for Code Generation}

\author{
Lize Shao \\
University of Virginia \\
\texttt{zgr3et@virginia.edu}
\And
Zichen Xie \\
University of Virginia \\
\texttt{graysonxie@virginia.edu}
\And
Zhizhen Chen \\
University of Virginia \\
\texttt{bpm6gu@virginia.edu}
\AND
Zhe Zeng \\
University of Virginia \\
\texttt{zhez@virginia.edu}
\And
Wenxi Wang \\
University of Virginia \\
\texttt{wenxiw@virginia.edu}
}

\hypersetup{
pdftitle={Self-Decomposing Diffusion for Code Generation},
pdfsubject={cs.LG, cs.CL, cs.SE},
pdfkeywords={Diffusion Language Models, Code Generation, Parallel Decoding, Planning},
}

\begin{document}
\maketitle

\begin{abstract}
Diffusion language models (DLMs) are emerging as a promising paradigm for code generation, offering the potential for fast, parallel decoding. However, generating complex programs remains challenging, as their interdependent components must maintain consistent interfaces, data flows, and shared assumptions. Aggressive parallel decoding can compromise these dependencies and undermine program correctness, while conservative decoding sacrifices efficiency. To address this trade-off, we propose planning program decomposition before implementation to coordinate interdependent units during parallel generation. We introduce Self-Decomposing Diffusion for Code (\ours{}), a novel framework that decomposes complex programs and implements their units in parallel, entirely through diffusion. \ours{} first generates a \emph{contract} specifying the program's decomposition and inter-unit dependencies, then uses it to guide coordinated parallel implementation. Across five benchmarks and two 7B diffusion backbones, \ours{} substantially improves the accuracy--latency trade-off, raising pass@1 by up to 31.1 percentage points while decoding 7.5--13.7$\times$ faster than the base models, and outperforming a fine-tuning baseline by up to 23.5 points with lower latency.

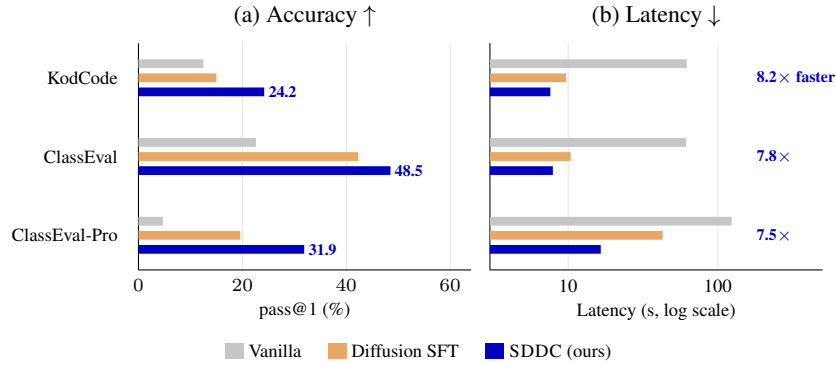
\begin{figure}[H]
    \providecommand{\teaseraxiswidth}{2.5cm}%
\providecommand{\teaseraxisheight}{2.7cm}
    \centering
    \pgfplotsset{
      teaser/.style={
        xbar, scale only axis, width=\teaseraxiswidth, height=\teaseraxisheight, bar width=3.3pt,
        symbolic y coords={ClassEval-Pro,ClassEval,KodCode}, ytick=data, enlarge y limits=0.22,
        y tick label style={font=\scriptsize},
        x tick label style={font=\scriptsize},
        axis lines*=left, xmajorgrids, grid style={gray!20},
        xtick style={draw=none}, ytick style={draw=none},
        label style={font=\scriptsize}, xlabel style={yshift=0.6ex},
        title style={font=\footnotesize, yshift=-0.8ex},
        clip=false,
      },
      tvan/.style={fill=gray!45, draw=none},
      tsft/.style={fill=orange!85!black!60, draw=none},
      tours/.style={fill=blue!75!black, draw=none},
    }
    \tikzset{tlab/.style={font=\tiny\bfseries, text=blue!75!black, inner sep=1.5pt}}
    \begin{tikzpicture}
    \begin{axis}[teaser, name=acc, xmin=0, xmax=64, xtick={0,20,40,60}, xlabel={pass@1 (\%)},
        title={(a) Accuracy $\uparrow$},
        legend to name=teaserlegend, legend columns=3, reverse legend,
        legend style={font=\scriptsize, draw=none, /tikz/every even column/.append style={column sep=8pt}},
        legend image code/.code={\draw[#1, draw=none] (0cm,-0.08cm) rectangle (0.24cm,0.12cm);}]
      \addplot[tours, nodes near coords align={horizontal}, nodes near coords style={tlab},
        nodes near coords={\pgfmathprintnumber[fixed, fixed zerofill, precision=1, assume math mode=true]{\pgfplotspointmeta}}]
        coordinates {(31.9,ClassEval-Pro) (48.5,ClassEval) (24.2,KodCode)};
      \addplot[tsft] coordinates {(19.6,ClassEval-Pro) (42.3,ClassEval) (15.0,KodCode)};
      \addplot[tvan] coordinates {(4.7,ClassEval-Pro) (22.6,ClassEval) (12.5,KodCode)};
      \legend{\ours{} (ours), Diffusion SFT, Vanilla}
    \end{axis}
    \begin{axis}[teaser, at={(acc.east)}, anchor=west, xshift=0.25cm, xmode=log, log origin=infty, xmin=3, xmax=500,
        xtick={10,100}, xticklabels={10,100}, yticklabels={}, xlabel={Latency (s, log scale)},
        title={(b) Latency $\downarrow$}]
      \addplot[tours] coordinates {(16.5,ClassEval-Pro) (7.9,ClassEval) (7.6,KodCode)};
      \addplot[tsft] coordinates {(42.8,ClassEval-Pro) (10.4,ClassEval) (9.7,KodCode)};
      \addplot[tvan] coordinates {(123.6,ClassEval-Pro) (61.3,ClassEval) (61.9,KodCode)};
      \node[tlab, anchor=west] at (axis cs:170,KodCode) {8.2$\times$ faster};
      \node[tlab, anchor=west] at (axis cs:170,ClassEval) {7.8$\times$};
      \node[tlab, anchor=west] at (axis cs:170,ClassEval-Pro) {7.5$\times$};
    \end{axis}
    \end{tikzpicture}

    \vspace{0.2ex}
    \pgfplotslegendfromname{teaserlegend}
    \caption{\textbf{\ours{} is more accurate and faster.} Dream-Coder 7B on the program-level benchmarks of
    Table~\ref{tab:program}: (a)~pass@1 (test pass rate on ClassEval and ClassEval-Pro); (b)~latency per problem and
    speedup over Vanilla.}
    \label{fig:teaser}
\end{figure}
\end{abstract}

\section{Introduction}
\label{sec:intro}

\ifdefined\arxivversion\else
\begin{figure}[t]
    \providecommand{\teaseraxiswidth}{2.5cm}%
\providecommand{\teaseraxisheight}{2.7cm}
    \centering
    \pgfplotsset{
      teaser/.style={
        xbar, scale only axis, width=\teaseraxiswidth, height=\teaseraxisheight, bar width=3.3pt,
        symbolic y coords={ClassEval-Pro,ClassEval,KodCode}, ytick=data, enlarge y limits=0.22,
        y tick label style={font=\scriptsize},
        x tick label style={font=\scriptsize},
        axis lines*=left, xmajorgrids, grid style={gray!20},
        xtick style={draw=none}, ytick style={draw=none},
        label style={font=\scriptsize}, xlabel style={yshift=0.6ex},
        title style={font=\footnotesize, yshift=-0.8ex},
        clip=false,
      },
      tvan/.style={fill=gray!45, draw=none},
      tsft/.style={fill=orange!85!black!60, draw=none},
      tours/.style={fill=blue!75!black, draw=none},
    }
    \tikzset{tlab/.style={font=\tiny\bfseries, text=blue!75!black, inner sep=1.5pt}}
    \begin{tikzpicture}
    \begin{axis}[teaser, name=acc, xmin=0, xmax=64, xtick={0,20,40,60}, xlabel={pass@1 (\%)},
        title={(a) Accuracy $\uparrow$},
        legend to name=teaserlegend, legend columns=3, reverse legend,
        legend style={font=\scriptsize, draw=none, /tikz/every even column/.append style={column sep=8pt}},
        legend image code/.code={\draw[#1, draw=none] (0cm,-0.08cm) rectangle (0.24cm,0.12cm);}]
      \addplot[tours, nodes near coords align={horizontal}, nodes near coords style={tlab},
        nodes near coords={\pgfmathprintnumber[fixed, fixed zerofill, precision=1, assume math mode=true]{\pgfplotspointmeta}}]
        coordinates {(31.9,ClassEval-Pro) (48.5,ClassEval) (24.2,KodCode)};
      \addplot[tsft] coordinates {(19.6,ClassEval-Pro) (42.3,ClassEval) (15.0,KodCode)};
      \addplot[tvan] coordinates {(4.7,ClassEval-Pro) (22.6,ClassEval) (12.5,KodCode)};
      \legend{\ours{} (ours), Diffusion SFT, Vanilla}
    \end{axis}
    \begin{axis}[teaser, at={(acc.east)}, anchor=west, xshift=0.25cm, xmode=log, log origin=infty, xmin=3, xmax=500,
        xtick={10,100}, xticklabels={10,100}, yticklabels={}, xlabel={Latency (s, log scale)},
        title={(b) Latency $\downarrow$}]
      \addplot[tours] coordinates {(16.5,ClassEval-Pro) (7.9,ClassEval) (7.6,KodCode)};
      \addplot[tsft] coordinates {(42.8,ClassEval-Pro) (10.4,ClassEval) (9.7,KodCode)};
      \addplot[tvan] coordinates {(123.6,ClassEval-Pro) (61.3,ClassEval) (61.9,KodCode)};
      \node[tlab, anchor=west] at (axis cs:170,KodCode) {8.2$\times$ faster};
      \node[tlab, anchor=west] at (axis cs:170,ClassEval) {7.8$\times$};
      \node[tlab, anchor=west] at (axis cs:170,ClassEval-Pro) {7.5$\times$};
    \end{axis}
    \end{tikzpicture}

    \vspace{0.2ex}
    \pgfplotslegendfromname{teaserlegend}
    \caption{\textbf{\ours{} is more accurate and faster.} Dream-Coder 7B on the program-level benchmarks of
    Table~\ref{tab:program}: (a)~pass@1 (test pass rate on ClassEval and ClassEval-Pro); (b)~latency per problem and
    speedup over Vanilla.}
    \label{fig:teaser}
    \vspace{-10pt}
    \end{figure}
\fi

\begin{figure*}[t]
\centering
\includegraphics[width=\textwidth]{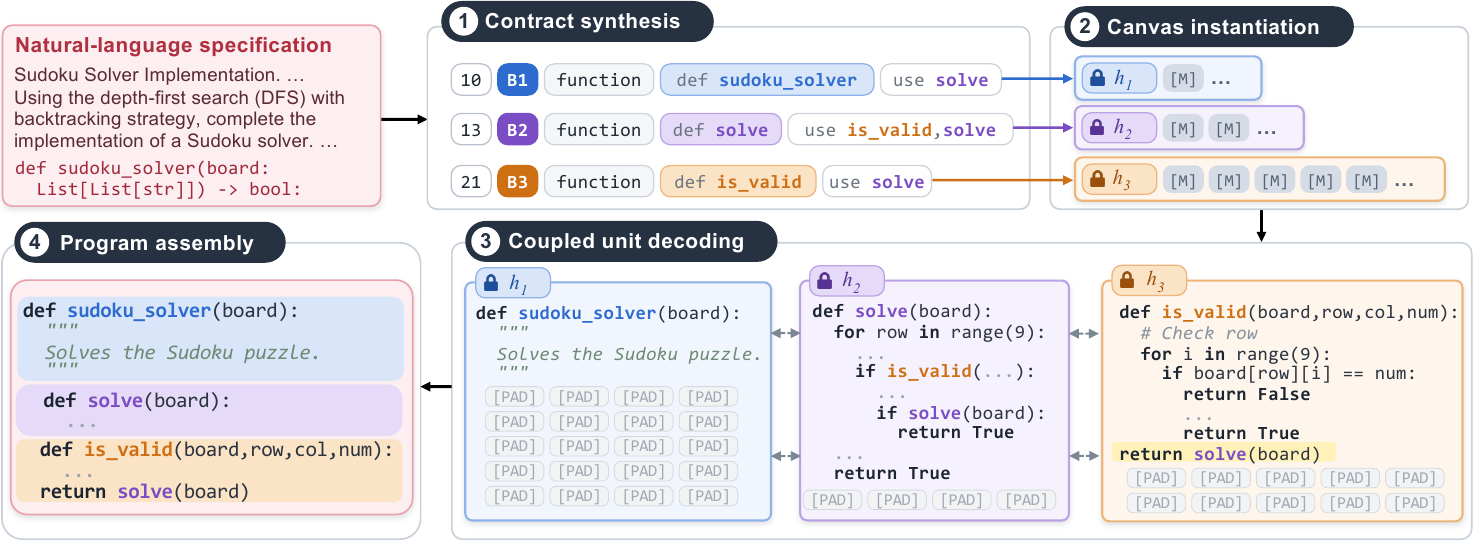}
\caption{Overview of \ours{} on a held-out KodCode task (abridged and adapted from the model's output). \textbf{(1)}~Contract synthesis denoises entries that give a budget, an index, a role, and the identifiers that each unit defines (\texttt{def}) and requires (\texttt{use}). \textbf{(2)}~Canvas instantiation fixes each entry's header $h_k$ at the start of its canvas. \textbf{(3)}~Coupled unit decoding denoises all canvases jointly; they also coordinate a decision that the contract leaves open, nesting both helpers inside \texttt{sudoku\_solver}, whose final \texttt{return} the third canvas writes (highlighted). \textbf{(4)}~Program assembly removes headers and padding and concatenates the units.}
\label{fig:overview}
\vspace{-10pt}
\end{figure*}

Language models are proficient at function-level programming, yet
generating complete programs remains difficult: the methods, helper
functions, and class headers of a program must be mutually consistent in
their identifiers, interfaces, and shared state, and an error in any
single unit invalidates the entire program. Models therefore perform far worse on class-level than on
function-level benchmarks~\citep{du2024classeval,rahman2025beyond,liang2026scaling},
and the strongest model evaluated on ClassEval-Pro completes fewer than
half of its classes~\citep{chen2026classevalpro}. 

A standard approach to such tasks, analogous to human software
development, is to decompose a problem into subproblems and to plan
their implementation before writing code. Planning has accordingly been
widely adopted for code generation with autoregressive models, which
outline solution steps or subfunctions before implementing
them~\citep{jiang2024self,zelikman2023parsel,le2024codechain,chen2024divide,wang2025planning}
or plan dependent edits across a repository~\citep{bairi2024codeplan}.
Diffusion language models can commit multiple tokens in a single forward
pass~\citep{nie2026large,ye2025dream,xie2025dream,gong2026diffucoder},
yet planning with them remains largely unexplored. Such planning by diffusion language models is
especially valuable for code, whose units interact through explicit
interfaces. A plan that specifies these interfaces in advance guides the reasoning about
decomposition, improves consistency across units, and enables their
parallel generation. Current diffusion code models decode programs
without such a plan, so committing many tokens at once introduces
inconsistencies across units~\citep{kang2026parallelbench,kim2026dapd}.

Recent work combines planning and diffusion with autoregressive
decoding to accelerate generation.
Building on parallel generation of semantically independent
text spans with autoregressive models~\citep{ning2024skeleton,jin2025learning},
Planned Diffusion~\citep{israel2026planned} autoregressively
generates a plan of chunk topics and lengths, then denoises
the planned chunks in parallel, whereas
ReFusion~\citep{li2026refusion} selects slots by diffusion and fills them
autoregressively. However, these methods are not designed to coordinate
interdependent program units, and both retain an
autoregressive stage that limits end-to-end parallelism.

We introduce Self-Decomposing Diffusion for Code (\ours{}),
which aims to improve both accuracy and decoding latency in
complex code generation (Figure~\ref{fig:overview}).
Our approach combines two complementary ideas:
\emph{decomposition} enables parallel implementation by dividing
a program into interacting units, while a \emph{contract}
promotes consistency by specifying these units and the interfaces
between them. \ours{} first denoises the contract, then
implements all units in parallel, binding each unit to its
contract entry and coordinating them through cross-unit
attention. Both stages are performed entirely by diffusion;
to our knowledge, \ours{} is the first approach that plans
the generation of complex, multi-unit programs with a diffusion
model.

Across five program-level and self-invoking benchmarks and two 7B
diffusion backbones, \ours{} improves pass@1 over Vanilla by up to 31.1 points while decoding 7.5--13.7$\times$ faster (Figure~\ref{fig:teaser}), and outperforms the Diffusion SFT baseline by up to 23.5 points. Our contributions are as follows:
\begin{itemize}[noitemsep, topsep=2pt, parsep=0pt,
                partopsep=0pt, leftmargin=*]
\item \textbf{Self-decomposing diffusion.}
We formulate complex code generation as diffusion over an explicit
program decomposition: a contract specifying program units and
their interfaces, which the model learns to generate from
static-analysis supervision.

\item \textbf{Contract-bound coupled decoding.}
We use the generated contract to guide joint denoising of
interdependent program units, enabling parallel implementation
while improving accuracy and cross-unit consistency.

\item \textbf{Improved latency--quality trade-off.}
\ours{} advances the latency--quality frontier of diffusion code
generation, and ablations confirm the contribution of each of its components. 
\end{itemize}

\section{Related Work}
\label{sec:related}

Planning and decomposition with autoregressive models are well established for complex code generation~\citep{jiang2024self,le2024codechain,chen2024divide,islam2024mapcoder,bairi2024codeplan}. This paper focuses on diffusion models for complex code generation. 

\paragraph{Diffusion models for code.}
Code diffusion models have advanced through pretraining and
reinforcement
learning~\citep{xie2025dream,gong2026diffucoder,fan2026stable},
adaptive infilling and decoding schedules~\citep{wu2026dreamon,dong2025saber},
constrained
decoding~\citep{suresh2026dingo,mundler2026constrained,zhang2026lookahead,shao2026constrained},
and structural guidance from abstract syntax
trees~\citep{zeng2026treediff} or anchors~\citep{xue2026ancoder}. These
methods generate a program as a single sequence and are evaluated mainly
on function-level benchmarks such as HumanEval~\citep{chen2021evaluating}
and MBPP~\citep{austin2021program}, rather than on classes or other
complex, multi-unit programs. \ours{} instead synthesizes a program-level
contract and generates interdependent units concurrently, modeling
explicitly the identifiers that connect them.

\paragraph{Planning with diffusion.}
Recent work has applied planning to diffusion-based
generation. Planned Diffusion~\citep{israel2026planned} generates
a plan of topics and lengths autoregressively and denoises the
planned chunks in parallel, ReFusion~\citep{li2026refusion}
selects fixed-length slots by diffusion and fills them
autoregressively, and Plan-Verify-Fill~\citep{li2026plan}
decodes verified anchor tokens first; templates and learned
policies further shape the generation
order~\citep{xiong2026unveiling,lee2026unlocking,huang2026reinforcing,kim2026ta,xu2026scheduling}.
These methods primarily organize generation around text chunks or token positions, without explicitly planning how program components interact. Complex code, however, requires coordinating interdependent units according to program structure and dependencies. \ours{} addresses this challenge through a code-specific contract that guides parallel implementation, using diffusion for both planning and generation.
\section{Preliminaries}
\label{sec:prelim}

We study code generation from a natural-language specification $\vc$.
The model generates a program $\vx=(x^1,\ldots,x^L)\in\gV^L$ over a
vocabulary $\gV$ that contains a padding token $\pad$; a separate token
$\mask\notin\gV$ marks corrupted positions.

\paragraph{Masked diffusion language models.}
Masked diffusion language models learn to reconstruct sequences
corrupted by an absorbing-state masking
process~\citep{austin2021structured,sahoo2024simple,shi2024simplified}:
at noise level $t\in[0,1]$, each token of the program (but not of the
specification), is replaced by $\mask$ independently with probability
$t$. A Transformer denoiser $p_\theta(x^i\mid\vx_t,\vc)$ attends
bidirectionally over the corrupted sequence $\vx_t$ and is trained with
the reweighted masked-token
cross-entropy~\citep{sahoo2024simple,shi2024simplified}
\begin{equation}
\mathcal{L}_{\textsc{mdm}}
=\E_{t,\vx_t}\Bigl[\frac{1}{t}\sum_{i\in\mathcal{M}_t}
-\log p_\theta(x^i\mid\vx_t,\vc)\Bigr],
\label{eq:mdm}
\end{equation}
where $t\sim\mathcal{U}(0,1)$ and $\mathcal{M}_t$ is the set of masked
positions. At inference, the decoder starts from a fully masked canvas,
predicts all masked positions in each forward pass, and commits a subset
of the predictions, so several tokens can be committed per step and in
any order.

\paragraph{Parallel sampling.}
Let $\eta$ denote the observed context, comprising the specification and
the committed tokens, and let $p_\star(\cdot\mid\eta)$ be the target
conditional distribution over the masked tokens. Committing a set $Q$ of
masked positions, chosen before their values are sampled, draws the
tokens $X^Q=(X^i)_{i\in Q}$ independently from their marginals. Even
with exact marginals, this incurs the error
\begin{equation}
\begin{aligned}
&\KL\Bigl(p_\star(X^Q\mid\eta)\,\Big\|\prod_{i\in Q}p_\star(X^i\mid\eta)\Bigr)\\
&\quad=\operatorname{TC}_\star(X^Q\mid\eta),
\end{aligned}
\label{eq:tc}
\end{equation}
the total correlation~\citep{watanabe1960information} of the committed
tokens, that is, the sum of their marginal entropies minus their joint
entropy under $p_\star(\cdot\mid\eta)$. It vanishes if and only if the
committed tokens are conditionally independent given $\eta$, so
parallel decoding is accurate only when the tokens committed together
are nearly independent given what has been revealed~\citep{wu2026fast}.

\section{Motivation}
\label{sec:principle}

To motivate the design of \ours{}, we decompose the error of committing
several tokens in parallel. Equation~\eqref{eq:tc} quantifies this error
for exact marginals; to extend it to a learned denoiser and a program
decomposed into disjoint units, let $Q$ denote the positions committed
in one iteration, selected given the observed context $\eta$ before
sampling, let $Q_k$ be the nonempty subset of $Q$ belonging to unit $k$,
and let $p_{\theta,i}(\cdot\mid\eta)$ denote the denoiser's marginal
distribution at position $i$.

\begin{proposition}[Decomposition of parallel-sampling error]
\label{prop:step}
If the following divergences are finite, independent
sampling from the denoiser marginals satisfies
\begin{equation}
\begin{aligned}
&\KL\Bigl(p_\star(X^Q\mid\eta)\,\Big\|\prod_{i\in Q}p_{\theta,i}(X^i\mid\eta)\Bigr)\\
&\quad=\underbrace{\operatorname{TC}_\star\bigl((X^{Q_k})_k\mid\eta\bigr)}_{\text{R1: cross-unit dependence}}
+\underbrace{\textstyle\sum_k\operatorname{TC}_\star(X^{Q_k}\mid\eta)}_{\text{R2: within-unit dependence}}\\
&\qquad+\underbrace{\textstyle\sum_{i\in Q}\KL\bigl(p_\star(X^i\mid\eta)\,\|\,p_{\theta,i}(X^i\mid\eta)\bigr)}_{\text{R3: marginal approximation error}}.
\end{aligned}
\label{eq:error-decomposition}
\end{equation}
\end{proposition}

\noindent The proof is given in Appendix~\ref{app:proofs}. Each term is a
source of error in parallel code generation, and \ours{} mitigates each
with a dedicated mechanism:
\begin{itemize}[noitemsep, topsep=2pt, parsep=0pt,
                partopsep=0pt, leftmargin=*]
\item[\textbf{R1}] \textbf{Cross-unit dependence.}
Units must coordinate their interactions, yet parallel generation can
lead to incompatible decisions across units. A contract that specifies
the identifiers each unit defines and requires makes the key cross-unit
relationships explicit before implementation.
\item[\textbf{R2}] \textbf{Within-unit dependence.}
Confidence-gated commitment restricts additional commitments to
high-confidence positions. For exact marginals, each additional position
whose most probable token has probability at least $\gamma\geq\frac12$
adds at most $H_{\mathrm b}(\gamma)+(1-\gamma)\log(|\gV|-1)$ to an upper
bound on this term, where $H_{\mathrm b}$ is the binary entropy
(Appendix~\ref{app:proofs}).
\item[\textbf{R3}] \textbf{Marginal approximation error.}
Accurate predictions require unit-specific context and information from
other units. Contract-bound headers identify the unit being implemented,
and coupled attention lets predictions condition on the evolving
implementations of other units.
\end{itemize}

\section{\ours{}}
\label{sec:method}

\ours{} generates a program in four stages (Figure~\ref{fig:overview} and
Algorithm~\ref{alg:ours}): \textcircled{1} \emph{contract synthesis}
denoises a contract that decomposes the task into program units;
\textcircled{2} \emph{canvas instantiation} allocates one canvas per
unit; \textcircled{3} \emph{coupled unit decoding} denoises all canvases
jointly; and \textcircled{4} \emph{program assembly} removes headers and
padding and concatenates the units in contract order.

\subsection{Stage 1: Contract Synthesis}
\label{sec:stage1}
Contract synthesis plans the program before any unit is implemented. We
represent a program as a concatenation of $K$ units,
$\vx=\vx^{(1)}\circ\cdots\circ\vx^{(K)}$, each a contiguous region of
source code with a structural role, such as a class header with its
constructor, a method, or a function. A contract
$\vz=(e_1,\ldots,e_K)$ assigns each unit an entry
$e_k=(b_k,k,\rho_k,\mathcal{D}_k,\mathcal{R}_k)$ that comprises a size
budget $b_k$ in multiples of $\nu$ tokens, the unit index, a structural
role, the identifiers $\mathcal{D}_k$ that the unit defines, and the
identifiers $\mathcal{R}_k$ that it reads before binding them. For
example, the entry \texttt{13 B2 function | def solve | use is\_valid,solve}
in Figure~\ref{fig:overview} allocates 13 budget units to a function
that defines \texttt{solve} and calls \texttt{is\_valid} and,
recursively, \texttt{solve}. We write $\operatorname{ser}(\vz)$ for the
serialized contract and call the serialization $h_k$ of $e_k$ without
its budget the \emph{header} of unit $k$.

\paragraph{Nominal interfaces.}
Identifiers form the interface between units, since a unit can call
functions and access attributes defined by other units without observing
their implementations. By declaring these identifiers, the contract
fixes the decisions on which cross-unit dependence concentrates, which
mitigates R1, and leaves all other decisions to the units.

\paragraph{Decoding.}
The contract is generated on a canvas $\vy$ of $P$ positions that is
initialized to $\mask^P$ and appended to the specification. Contract
positions attend bidirectionally to one another and to $\vc$, so that
every entry can condition on tokens committed in any other entry. At
each masked position $j$, the decoder selects tokens from the admissible
set
\begin{equation}
\mathcal{A}_j(\vy)=
\begin{cases}
\gV\setminus\{\pad\} & y^{j+1}\neq\pad\\
\gV & \text{otherwise}
\end{cases}
\label{eq:admissible}
\end{equation}
with the convention $y^{P+1}=\pad$. Padding is thus admissible only at
the final position or immediately before committed padding, which keeps
the padding a contiguous suffix and prevents premature truncation. In
each iteration $n$, the decoder proposes the most probable admissible
token $\hat y^j$
at every masked position, with confidence $\kappa_j$ equal to its
probability, and commits all proposals with $\kappa_j\geq\gamma_z$, but
at least the $\lceil|\mathcal{M}(\vy)|/(S_z-n+1)\rceil$ most confident
ones, where $\mathcal{M}(\vy)$ denotes the masked positions of $\vy$;
this rule bounds synthesis by $S_z$ iterations. A parser finally
extracts the well-formed entries $\hat\vz$.

\subsection{Stage 2: Canvas Instantiation}
\label{sec:stage2}
Each entry $e_k$ receives a canvas of $\ell_k=\lambda b_k$ positions,
initialized as $\vw_0^{(k)}=h_k\circ\mask^{\ell_k-|h_k|}$; we write
$\Phi_\lambda(\vz)$ for the initialized canvases. The header stays
fixed during decoding and binds the canvas to its entry (R3), so the
denoiser need not infer from position which unit a canvas implements.
Budgets specify capacity rather than length: the scale $\lambda\geq\nu$
provides headroom, and unused positions are filled with $\pad$. Let
$\mathcal{Z}_\lambda$ be the set of nonempty contracts whose headers fit
their canvases, $|h_k|\leq\ell_k$. For any other parsed contract,
generation returns the failure outcome $\bot$, which we count as an
incorrect program without resampling.

\subsection{Stage 3: Coupled Unit Decoding}
\label{sec:stage3}
All canvases are denoised in parallel. The context is encoded once as
$\mathcal{H}=\mathsf{Enc}_\theta([\vc;\operatorname{ser}(\vz)])$, and
every canvas attends to $\mathcal{H}$ and to the current state of all
canvases, so that units condition on one another's partial
implementations, including decisions that the contract leaves open
(R3). In each iteration, every unfinished canvas samples a candidate
$a_{k,j}$ at each of its masked positions and commits the candidate with
the highest probability; it also commits every other masked position
whose maximum probability is at least $\gamma_u$ to its most probable
token (R2). The operator $\mathsf{F}$ then fills all masked positions
after the first committed $\pad$ of a canvas with $\pad$, and a canvas
is complete when no masks remain. As an ablation, \emph{isolated}
decoding lets each canvas attend only to $\mathcal{H}$ and to itself.

\begin{algorithm}[t]
\caption{Generation with \ours{}. $\mathcal{M}(\cdot)$ denotes the masked
positions of a canvas, $\operatorname{Top}_r(\kappa;M)$ the $r$ positions
in $M$ with the largest confidences $\kappa$, and $\Pi$ the token
distributions returned by one evaluation of the denoiser
$\mathsf{D}^{\mathrm{con}}_\theta$ or $\mathsf{D}^{\mathrm{unit}}_\theta$
over the masked contract or unit positions; $A_{\hat\vz}$ removes headers
and padding and concatenates the units. Other symbols are defined in
Sections~\ref{sec:stage1}--\ref{sec:stage3}.}
\label{alg:ours}
\small
\begin{algorithmic}[1]
\Require $\theta,\vc$; $P,S_z,\lambda\in\mathbb{N}_{>0}$;
         $\gamma_z,\gamma_u\in[0,1]$
\Ensure Program $\vx$ or failure outcome $\bot$
\State $\vy\gets\mask^P$;\ $n\gets0$ \Comment{contract synthesis}
\While{$\mathcal{M}(\vy)\neq\varnothing\ \land\ n<S_z$}
  \State $n\gets n+1$;\ $M\gets\mathcal{M}(\vy)$;\
         $\Pi\gets\mathsf{D}^{\mathrm{con}}_\theta(\vy;\vc)$
  \State $\hat y^j\gets\arg\max_{v\in\mathcal{A}_j(\vy)}\Pi_j(v)$;\
         $\kappa_j\gets\Pi_j(\hat y^j)$
  \State $r\gets\max\bigl(\lceil\frac{|M|}{S_z-n+1}\rceil,
         |\{j\in M:\kappa_j\geq\gamma_z\}|\bigr)$
  \State $y^j\gets\hat y^j$ \quad $(j\in\operatorname{Top}_r(\kappa;M))$
\EndWhile
\State $\hat\vz=(e_1,\ldots,e_K)\gets\operatorname{parse}(\vy)$
\State \textbf{if} $\hat\vz\notin\mathcal{Z}_\lambda$ \textbf{return} $\bot$
       \Comment{canvas instantiation}
\State $\mathbf{w}\gets\Phi_\lambda(\hat\vz)$;\
       $\mathcal{H}\gets\mathsf{Enc}_\theta([\vc;\operatorname{ser}(\hat\vz)])$
\While{$\exists k:\mathcal{M}(\vw^{(k)})\neq\varnothing$}
       \Comment{coupled unit decoding}
  \State $\Pi\gets\mathsf{D}^{\mathrm{unit}}_\theta(\mathbf{w};\mathcal{H})$
  \ForAll{$k:M_k=\mathcal{M}(\vw^{(k)})\neq\varnothing$}
    \Comment{in parallel}
    \State $a_{k,j}\sim\Pi_{k,j}$ \quad $(j\in M_k)$
    \State $j_k\gets\arg\max_{j\in M_k}\Pi_{k,j}(a_{k,j})$
    \State $w^{(k),j_k}\gets a_{k,j_k}$
    \State $U_k\gets\{j\in M_k\setminus\{j_k\}:
           \max_v\Pi_{k,j}(v)\geq\gamma_u\}$
    \State $w^{(k),j}\gets\arg\max_v\Pi_{k,j}(v)$ \quad $(j\in U_k)$
    \State $\vw^{(k)}\gets\mathsf{F}(\vw^{(k)})$
  \EndFor
\EndWhile
\State \Return $A_{\hat\vz}(\mathbf{w})$ \Comment{program assembly}
\end{algorithmic}
\end{algorithm}

\section{SDDC Training}
\label{sec:training}

For each training pair $(\vc,\vx)$, we extract a reference contract
$\vz^\star$ from $\vx$ and train two views of the example: a contract
view for synthesis and a unit view for implementation.

\paragraph{Contract extraction.}
Decomposition supervision comes from static analysis of reference
programs. We segment $\vx$ at the definition boundaries of its abstract
syntax tree into class headers with their constructors, methods, and
top-level functions, attaching module-level statements to adjacent
segments. Each segment forms its own unit by default, and adjacent
segments are merged by annotation for a curated subset
(Section~\ref{sec:exp-setup}). Name analysis then yields $\mathcal{D}_k$
and $\mathcal{R}_k$, the role $\rho_k$ follows from the syntactic
category, and the budget $b_k$ covers the header and the unit with a
small slack, with $\nu=10$ (Appendix~\ref{app:data}). The analyzer only
constructs training targets: at inference, the model generates the
contract from the specification alone.

\paragraph{Contract view.}
The target is the padded serialization
$\vy^\star=\operatorname{ser}(\vz^\star)\circ\pad^{\,P-|\operatorname{ser}(\vz^\star)|}$,
which is corrupted as in Section~\ref{sec:prelim} while $\vc$ remains
intact. With bidirectional attention over the contract canvas, the loss
is
\begin{equation}
\mathcal{L}_{\textsc{con}}
=\E\Bigl[\frac{1}{t}\sum_{j\in\mathcal{M}_t}
-\log p_\theta(y^{\star j}\mid\vy_t,\vc)\Bigr],
\label{eq:loss-con}
\end{equation}
where the expectation is over $t\sim\mathcal{U}(0,1)$ and the corrupted
canvas $\vy_t$, and $\mathcal{M}_t$ denotes its masked positions. Supervising the padding teaches the model how many units to
create; contracts longer than $P$ positions are excluded from this view.

\paragraph{Unit view.}
Each unit forms a canvas target $h_k\circ\vx^{(k)}\circ\pad^{\,r_k}$
with a padding surplus $r_k$. Because \ours{} allocates canvases in
proportion to predicted budgets, unused capacity grows with the unit, so
we draw each surplus independently from
\begin{equation}
\begin{aligned}
r_k\sim{}&\tfrac12\,\mathcal{U}\{0,\ldots,10\}\\
&+\tfrac12\,\mathcal{U}\{0,\ldots,\max(10,|\vx^{(k)}|)\},
\end{aligned}
\label{eq:surplus}
\end{equation}
and concatenate the canvases into $\vw$. Scaling the surplus with the
unit length teaches the model to pad unused capacity instead of
extending the unit into the next one (R3), which keeps oversized
canvases safe at inference. Headers are
never masked or supervised, and all other canvas positions are masked
independently with probability $t$, which yields
\begin{equation}
\begin{aligned}
\mathcal{L}_{\textsc{unit}}
={}&\E\Bigl[\frac{1}{t}\sum_{j\in\mathcal{M}_t}\\
&-\log p_\theta(w^j\mid\vw_t,\vz^\star,\vc)\Bigr],
\end{aligned}
\label{eq:loss-unit}
\end{equation}
where the expectation also covers the surpluses $r_k$, and
$\mathcal{M}_t$ denotes the masked canvas positions. The context
$[\vc;\operatorname{ser}(\vz^\star)]$ is encoded with causal attention,
and canvas positions attend to the context and to one another, or only
to themselves in the isolated variant. The causal context also admits an
auxiliary next-token loss $\mathcal{L}_{\textsc{ctx}}$ on its contract
tokens; \ours{} does not use it for generation, but it allows the same
weights to write contracts left to right (Section~\ref{sec:exp-analysis}).

\paragraph{Objective.}
Both views train the same parameters with
$\mathcal{L}=\E_{(\vc,\vx)}[\mathcal{L}_{\textsc{con}}+\mathcal{L}_{\textsc{unit}}+\mathcal{L}_{\textsc{ctx}}]$.
The unit view conditions on the reference contract during training and
on the generated contract at inference.

\section{Experiments}
\label{sec:experiments}

We evaluate the latency--quality trade-off of \ours{}
(Section~\ref{sec:main-results}) and analyze how its mechanisms mitigate
the error sources identified in Section~\ref{sec:principle}
(Section~\ref{sec:exp-analysis}).

\begin{table*}[t]
    \caption{Program-level composition: pass@1 and pass@5 (\%) on KodCode (held-out), ClassEval, and ClassEval-Pro. Class: classes that pass all of their tests; Test: unit tests (ClassEval) or test classes (ClassEval-Pro) passed. \textcolor{blue!70!black}{Blue} rows give the mean latency per problem in seconds and the speedup over Vanilla; pass@5 latency covers five samples. Values next to the scores give the change from Vanilla (\hlg{green} $\geq$10 points, \hlly{yellow} $<$10 points, \hlr{red} a decrease). pass@5 samples at temperature 0.4 and pass@1 at 0.2 (Appendix~\ref{app:eval}). Vanilla uses 512 iterations and Diffusion SFT threshold decoding (256 iterations for DiffuCoder on KodCode). The best score per backbone is \textbf{bolded}.}
    \label{tab:program}
    \centering
    \renewcommand{\lat}[1]{\textcolor{blue!70!black}{#1}}%
    \renewcommand{\arraystretch}{1.05}
    \setlength{\tabcolsep}{4pt}
    \resizebox{\textwidth}{!}{%
    \begin{tabular}{@{}ll ll llll llll@{}}
    \toprule
    & & \multicolumn{2}{l}{\textbf{KodCode (held-out)}} & \multicolumn{4}{l}{\textbf{ClassEval}} & \multicolumn{4}{l}{\textbf{ClassEval-Pro}} \\
    \cmidrule(lr){3-4} \cmidrule(lr){5-8} \cmidrule(lr){9-12}
    & & & & \multicolumn{2}{l}{pass@1} & \multicolumn{2}{l}{pass@5} & \multicolumn{2}{l}{pass@1} & \multicolumn{2}{l}{pass@5} \\
    \cmidrule(lr){5-6} \cmidrule(lr){7-8} \cmidrule(lr){9-10} \cmidrule(lr){11-12}
    \textbf{Model} & \textbf{Method} & pass@1 & pass@5 & Class & Test & Class & Test & Class & Test & Class & Test \\
    \midrule
    \multirow{6}{*}{Dream-Coder 7B}
      & Vanilla              & 12.5 & 6.7 & 9.0 & 22.6 & 6.0 & 23.2 & 3.3 & 4.7 & 4.0 & 6.8 \\
      &                      & \lat{61.9} & \lat{308.0} & \lat{61.3} & \lat{61.3} & \lat{306.9} & \lat{306.9} & \lat{123.6} & \lat{123.6} & \lat{619.7} & \lat{619.7} \\
      & Diffusion SFT        & 15.0 \hlly{(+2.5)} & 18.3 \hlg{(+11.6)} & 7.0 \hlr{($-$2.0)} & 42.3 \hlg{(+19.7)} & 8.0 \hlly{(+2.0)} & 48.3 \hlg{(+25.1)} & 1.3 \hlr{($-$2.0)} & 19.6 \hlg{(+14.9)} & 3.0 \hlr{($-$1.0)} & 32.0 \hlg{(+25.2)} \\
      &                      & \lat{9.7 \spd{6.4}} & \lat{51.3 \spd{6.0}} & \lat{10.4 \spd{5.9}} & \lat{10.4 \spd{5.9}} & \lat{55.5 \spd{5.5}} & \lat{55.5 \spd{5.5}} & \lat{42.8 \spd{2.9}} & \lat{42.8 \spd{2.9}} & \lat{220.8 \spd{2.8}} & \lat{220.8 \spd{2.8}} \\
      & \textbf{\ours{} (Ours)} & \textbf{24.2} \hlg{(+11.7)} & \textbf{31.2} \hlg{(+24.5)} & \textbf{14.0} \hlly{(+5.0)} & \textbf{48.5} \hlg{(+25.9)} & \textbf{18.0} \hlg{(+12.0)} & \textbf{63.2} \hlg{(+40.0)} & \textbf{3.7} \hlly{(+0.4)} & \textbf{31.9} \hlg{(+27.2)} & \textbf{4.3} \hlly{(+0.3)} & \textbf{46.1} \hlg{(+39.3)} \\
      &                      & \lat{7.6 \spd{8.2}} & \lat{36.9 \spd{8.3}} & \lat{7.9 \spd{7.8}} & \lat{7.9 \spd{7.8}} & \lat{39.5 \spd{7.8}} & \lat{39.5 \spd{7.8}} & \lat{16.5 \spd{7.5}} & \lat{16.5 \spd{7.5}} & \lat{81.0 \spd{7.7}} & \lat{81.0 \spd{7.7}} \\
    \cmidrule(l){2-12}
    \multirow{6}{*}{DiffuCoder 7B}
      & Vanilla              & 7.9 & 8.8 & 6.0 & 16.1 & 3.0 & 17.0 & \textbf{4.0} & 6.0 & 4.0 & 9.2 \\
      &                      & \lat{61.8} & \lat{308.1} & \lat{61.5} & \lat{61.5} & \lat{307.0} & \lat{307.0} & \lat{124.1} & \lat{124.1} & \lat{620.2} & \lat{620.2} \\
      & Diffusion SFT        & 10.8 \hlly{(+2.9)} & 13.3 \hlly{(+4.5)} & 4.0 \hlr{($-$2.0)} & 25.6 \hlly{(+9.5)} & 4.0 \hlly{(+1.0)} & 27.1 \hlg{(+10.1)} & 0.0 \hlr{($-$4.0)} & 4.8 \hlr{($-$1.2)} & 0.0 \hlr{($-$4.0)} & 5.1 \hlr{($-$4.1)} \\
      &                      & \lat{30.7 \spd{2.0}} & \lat{153.5 \spd{2.0}} & \lat{13.4 \spd{4.6}} & \lat{13.4 \spd{4.6}} & \lat{69.9 \spd{4.4}} & \lat{69.9 \spd{4.4}} & \lat{32.1 \spd{3.9}} & \lat{32.1 \spd{3.9}} & \lat{165.9 \spd{3.7}} & \lat{165.9 \spd{3.7}} \\
      & \textbf{\ours{} (Ours)} & \textbf{23.3} \hlg{(+15.4)} & \textbf{24.2} \hlg{(+15.4)} & \textbf{13.0} \hlly{(+7.0)} & \textbf{47.2} \hlg{(+31.1)} & \textbf{20.0} \hlg{(+17.0)} & \textbf{64.0} \hlg{(+47.0)} & 3.0 \hlr{($-$1.0)} & \textbf{28.3} \hlg{(+22.3)} & \textbf{4.7} \hlly{(+0.7)} & \textbf{42.7} \hlg{(+33.5)} \\
      &                      & \lat{5.3 \spd{11.7}} & \lat{23.6 \spd{13.1}} & \lat{4.5 \spd{13.7}} & \lat{4.5 \spd{13.7}} & \lat{22.1 \spd{13.9}} & \lat{22.1 \spd{13.9}} & \lat{13.5 \spd{9.2}} & \lat{13.5 \spd{9.2}} & \lat{62.4 \spd{9.9}} & \lat{62.4 \spd{9.9}} \\
    \bottomrule
    \end{tabular}}
\end{table*}

\begin{table}[t]
    \caption{Self-invoking problem solving (pass@1 and pass@5, \%), formatted as in Table~\ref{tab:program}. \textcolor{blue!70!black}{Blue} rows give the mean latency per problem in seconds and the speedup over Vanilla. Values next to the scores give the change from Vanilla (\hlg{green} $\geq$10 points, \hlly{yellow} $<$10 points, \hlr{red} a decrease). Diffusion SFT uses threshold decoding (256 iterations for DiffuCoder on HumanEval Pro); Table~\ref{tab:sft-sweep} adds further configurations.}
    \label{tab:selfinvoking}
    \centering
    \renewcommand{\lat}[1]{\textcolor{blue!70!black}{#1}}%
    \renewcommand{\arraystretch}{1.05}
    \setlength{\tabcolsep}{3pt}
    \providecommand{\narrowtablewidth}{\columnwidth}%
    \resizebox{\narrowtablewidth}{!}{%
    \begin{tabular}{@{}ll llll@{}}
    \toprule
    & & \multicolumn{2}{l}{\textbf{HumanEval Pro}} & \multicolumn{2}{l}{\textbf{MBPP Pro}} \\
    \cmidrule(lr){3-4} \cmidrule(lr){5-6}
    \textbf{Model} & \textbf{Method} & pass@1 & pass@5 & pass@1 & pass@5 \\
    \midrule
    \multirow{6}{*}{\shortstack[l]{Dream-\\Coder 7B}}
      & Vanilla              & 40.2 & 19.5 & 51.9 & 33.9 \\
      &                      & \lat{39.4} & \lat{197.3} & \lat{34.7} & \lat{173.4} \\
      & Diffusion SFT        & 26.8 \hlr{($-$13.4)} & 31.7 \hlg{(+12.2)} & 36.0 \hlr{($-$15.9)} & 38.9 \hlly{(+5.0)} \\
      &                      & \lat{4.2 \spd{9.4}} & \lat{21.7 \spd{9.1}} & \lat{2.3 \spd{14.9}} & \lat{12.2 \spd{14.2}} \\
      & \textbf{\ours{} (Ours)} & \textbf{44.5} \hlly{(+4.3)} & \textbf{42.1} \hlg{(+22.6)} & \textbf{52.6} \hlly{(+0.7)} & \textbf{56.1} \hlg{(+22.2)} \\
      &                      & \lat{5.1 \spd{7.7}} & \lat{25.6 \spd{7.7}} & \lat{4.4 \spd{8.0}} & \lat{21.6 \spd{8.0}} \\
    \cmidrule(l){2-6}
    \multirow{6}{*}{\shortstack[l]{Diffu-\\Coder 7B}}
      & Vanilla              & 31.1 & 28.0 & 48.1 & 42.1 \\
      &                      & \lat{39.5} & \lat{197.5} & \lat{34.7} & \lat{173.5} \\
      & Diffusion SFT        & 20.1 \hlr{($-$11.0)} & 17.7 \hlr{($-$10.3)} & 31.5 \hlr{($-$16.6)} & 36.2 \hlr{($-$5.9)} \\
      &                      & \lat{19.7 \spd{2.0}} & \lat{98.8 \spd{2.0}} & \lat{2.8 \spd{12.6}} & \lat{14.1 \spd{12.3}} \\
      & \textbf{\ours{} (Ours)} & \textbf{40.9} \hlly{(+9.8)} & \textbf{51.2} \hlg{(+23.2)} & \textbf{49.7} \hlly{(+1.6)} & \textbf{54.0} \hlg{(+11.9)} \\
      &                      & \lat{3.2 \spd{12.4}} & \lat{15.0 \spd{13.1}} & \lat{3.2 \spd{10.8}} & \lat{15.6 \spd{11.1}} \\
    \bottomrule
    \end{tabular} }
\end{table}

\begin{figure*}[t]
\centering
\pgfplotsset{
  frontier/.style={
    scale only axis, width=0.27\textwidth, height=2.6cm,
    xmin=4, xmax=80,
    xtick={5,10,20,40,80}, xticklabels={5,10,20,40,80},
    ymin=0, ymax=28, ytick={0,5,10,15,20,25},
    xlabel={Latency per problem (s, log scale)},
    ylabel={pass@1 (\%)},
    grid=major, grid style={gray!20},
    tick label style={font=\scriptsize},
    label style={font=\footnotesize},
    title style={font=\footnotesize, yshift=-1ex},
    clip=false,
  },
  oursdef/.style={only marks, mark=star, mark size=4pt, line width=1.2pt, color=blue!75!black},
  oursalt/.style={only marks, mark=o, mark size=2.2pt, line width=0.8pt, color=blue!75!black},
  sftfix/.style={dashed, line width=0.8pt, mark=square*, mark size=1.7pt, mark options={solid}, color=orange!85!black},
  sftthr/.style={only marks, mark=diamond*, mark size=3pt, color=orange!85!black},
  vanfix/.style={dashed, line width=0.8pt, mark=triangle*, mark size=2.2pt, mark options={solid}, color=gray},
  vanthr/.style={only marks, mark=diamond, mark size=3pt, line width=0.8pt, color=gray},
}
\tikzset{itlab/.style={font=\tiny, inner sep=1.5pt}}
\begin{tikzpicture}
\begin{axis}[frontier, xmode=log, title={Dream-Coder 7B}]
  \addplot[sftfix] coordinates {(7.68,5.0) (15.36,7.5) (30.76,9.6)};
  \addplot[sftthr] coordinates {(9.72,15.0)};
  \addplot[vanfix] coordinates {(15.40,1.2) (30.9,4.2) (61.9,12.5)};
  \addplot[oursalt] coordinates {(8.56,21.7) (9.81,22.9) (12.46,25.8)};
  \addplot[oursdef] coordinates {(7.55,24.2)};
  \node[itlab, anchor=south, text=orange!85!black] at (axis cs:7.68,5.0) {64};
  \node[itlab, anchor=south, text=orange!85!black] at (axis cs:15.36,7.5) {128};
  \node[itlab, anchor=south, text=orange!85!black] at (axis cs:30.76,9.6) {256};
  \node[itlab, anchor=north, text=gray] at (axis cs:15.40,1.2) {128};
  \node[itlab, anchor=north, text=gray] at (axis cs:30.9,4.2) {256};
  \node[itlab, anchor=north, text=gray] at (axis cs:61.9,12.5) {512};
\end{axis}
\end{tikzpicture}\hspace{4mm}%
\begin{tikzpicture}
\begin{axis}[frontier, xmode=log, title={DiffuCoder 7B},
  ylabel={}, yticklabels={},
  legend style={font=\scriptsize, draw=none, at={(1.04,0.5)}, anchor=west,
                /tikz/every even row/.append style={row sep=1pt}},
  legend cell align=left]
  \addplot[sftfix] coordinates {(7.70,6.2) (15.39,7.9) (30.73,10.8)};
  \addlegendentry{Diffusion SFT, fixed iterations}
  \addplot[sftthr] coordinates {(11.63,10.0)};
  \addlegendentry{Diffusion SFT, threshold}
  \addplot[vanfix] coordinates {(15.36,2.9) (30.74,6.7) (61.78,7.9)};
  \addlegendentry{Vanilla, fixed iterations}
  \addplot[vanthr] coordinates {(6.79,4.2)};
  \addlegendentry{Vanilla, threshold}
  \addplot[oursalt] coordinates {(8.90,22.5) (9.55,21.7) (12.37,22.1)};
  \addlegendentry{\ours{}, other contract schedules}
  \addplot[oursdef] coordinates {(5.26,23.3)};
  \addlegendentry{\ours{} (default)}
  \node[itlab, anchor=south, text=orange!85!black] at (axis cs:7.70,6.2) {64};
  \node[itlab, anchor=south, text=orange!85!black] at (axis cs:15.39,7.9) {128};
  \node[itlab, anchor=south, text=orange!85!black] at (axis cs:30.73,10.8) {256};
  \node[itlab, anchor=north, text=gray] at (axis cs:15.36,2.9) {128};
  \node[itlab, anchor=north, text=gray] at (axis cs:30.74,6.7) {256};
  \node[itlab, anchor=north, text=gray] at (axis cs:61.78,7.9) {512};
\end{axis}
\end{tikzpicture}
\caption{Latency--quality trade-off on KodCode (held-out). Each point is one decoding configuration, labeled with its number of iterations; the upper left is better. On both backbones, the default configuration of \ours{} (star) dominates every baseline configuration, and its variants with other contract schedules and single commitment (circles) lie above all of them. Threshold decoding of the Dream-Coder base model (0.8 pass@1, not timed) is omitted. }
\label{fig:frontier}
\vspace{-10pt}
\end{figure*}
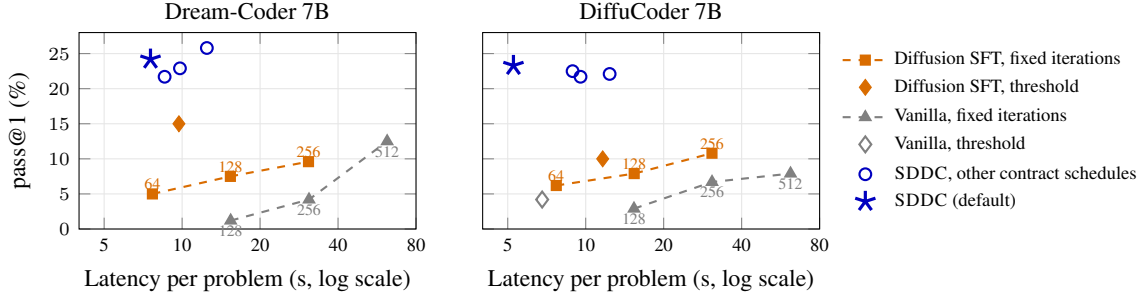

\begin{table}[!t]
    \caption{Binding (headers and length-scaled padding) and coupled attention on KodCode (curated subset, left-to-right contracts, $\lambda=10$; held-out). Misbound: canvases whose first definition belongs to another entry; Overrun: canvases that continue into the next unit (\%).}
    \label{tab:analysis}
    \centering
    \footnotesize
    \setlength{\tabcolsep}{4pt}
    \begin{tabular}{@{}ll ccc@{}}
    \toprule
    Attention & Binding & pass@1 & Misbound & Overrun \\
    \midrule
    isolated &            & 4.2  & 32.3 & 10.1 \\
    coupled  &            & 11.2 & 11.0 & 4.9 \\
    isolated & \checkmark & 13.3 & 0.1  & 1.0 \\
    coupled  & \checkmark & \textbf{14.2} & \textbf{0.0} & \textbf{0.9} \\
    \bottomrule
    \end{tabular}
\end{table}

\subsection{Experimental Setup}
\label{sec:exp-setup}

\paragraph{Models and training.}
We fine-tune two open 7B masked diffusion code models,
Dream-Coder-v0-Instruct-7B~\citep{xie2025dream} and
DiffuCoder-7B-Instruct~\citep{gong2026diffucoder}. The training corpus
contains 15,973 problems from KodCode-V1~\citep{xu2025kodcode} whose
verified solutions define at least three functions or four methods. For
2,770 of these programs, an LLM annotator merged adjacent units
following a written guideline; the corpus yields 25.5K unit-view and
25.1K contract-view examples. Each backbone is trained for six epochs, about
13 hours on four GPUs (Appendix~\ref{app:impl}).

\paragraph{Decoding.}
\ours{} uses $P=224$, $\gamma_z=0.95$, $S_z=112$, $\lambda=14$, and
$\gamma_u=0.99$, sampling with temperature 0.2 and top-$p$ 0.95
(Appendix~\ref{app:decoding}). All decoding hyperparameters
are tuned on KodCode; for ClassEval-Pro, whose specifications list up to
27 methods, we enlarge the contract canvas to $P=320$.

\paragraph{Benchmarks.}
We evaluate on five benchmarks in two groups. \emph{Program-level
composition} requires complete programs whose units interact through
shared state and calls. \emph{KodCode (held-out)} contains 240 problems
withheld from training and stratified by structure and difficulty,
scored by their pytest suites. \emph{ClassEval}~\citep{du2024classeval}
contains 100 hand-written classes whose skeletons specify interdependent
methods. \emph{ClassEval-Pro}~\citep{chen2026classevalpro} contains 300
cross-domain classes specified only in prose, most of which exceed our
2,048-token training length. On both class benchmarks, we report the
class pass rate, which requires all tests of a class to pass, and the
test pass rate, which credits partially correct classes
(Appendix~\ref{app:eval}). \emph{Self-invoking problem solving}
isolates a single cross-unit dependency: HumanEval Pro and MBPP
Pro~\citep{yu2025humaneval} extend HumanEval~\citep{chen2021evaluating}
and MBPP~\citep{austin2021program} with 164 and 378 problems in which
the solution to a base problem must be invoked to solve a harder one.

\paragraph{Baselines.}
\emph{Vanilla} decodes each released backbone over a full canvas with
512 iterations, its most accurate setting on every benchmark.
\emph{Diffusion SFT} fine-tunes the same backbone on exactly the same
programs with the same recipe and the standard masked diffusion
objective, which isolates the effect of the generation scheme from that
of the training data; we report the more accurate of
confidence-threshold decoding~\citep{wu2026fast} and 256 iterations on
each benchmark. Appendix~\ref{app:decoding} lists all evaluated
configurations.

\paragraph{Metrics.}
We report pass@1 at temperature 0.2 and pass@5 from five samples at
temperature 0.4. Latency is the mean wall-clock time per problem on one
NVIDIA RTX PRO 6000 GPU with batch size one, including contract
synthesis; for pass@5, it covers all five samples.

\subsection{Main Results}
\label{sec:main-results}

\paragraph{Program-level composition.}
Table~\ref{tab:program} shows that \ours{} is the fastest method on all
three program-level benchmarks with both backbones and the most accurate
on all but one metric. On KodCode, it attains 24.2 and 23.3 pass@1,
compared with 15.0 and 10.8 for Diffusion SFT and 12.5 and 7.9 for the
base models, while answering in 7.6 and 5.3 seconds, 8.2$\times$ and
11.7$\times$ faster than the base models. On ClassEval, it raises the
class pass rate to 14.0 and 13.0, from at most 9.0 and 6.0, and passes
48.5\% and 47.2\% of unit tests, compared with at most 42.3\% and 25.6\%.
The advantage widens on ClassEval-Pro, the most compositional benchmark,
where \ours{} passes 31.9\% and 28.3\% of test classes, compared with at
most 19.6\% and 6.0\%. The only exception, the DiffuCoder base model on
complete ClassEval-Pro classes (4.0 vs.\ 3.0), amounts to three of 300
classes at 9.2$\times$ the latency. Diffusion SFT on the
same programs, in contrast, lowers the class pass rate on both class
benchmarks with both backbones, so the gains of \ours{} stem from
generating programs through contracts rather than from the training
data.

\paragraph{Self-invoking problem solving.}
\ours{} also attains the highest pass@1 on HumanEval Pro and MBPP Pro
with both backbones (Table~\ref{tab:selfinvoking}). The base models
approach this accuracy only with 512 iterations, at 35--40 seconds per
problem; \ours{} exceeds them by up to 9.8 points while running
7.7--12.4$\times$ faster. Diffusion SFT trails the base models by
11.0--16.6 points; even with 512 iterations, it surpasses \ours{} only on
MBPP Pro with Dream-Coder (54.2 vs.\ 52.6), at about eight times the
latency (Table~\ref{tab:sft-sweep}). Since each
problem couples a base function with a function that must invoke it,
these results indicate that the dependency declared in the contract is
honored even though the two units are generated in parallel.

\paragraph{Sampling multiple programs.}
With five samples, \ours{} is the most accurate method in every pass@5
column of Tables~\ref{tab:program} and~\ref{tab:selfinvoking}; on
ClassEval, it passes 63.2\% and 64.0\% of unit tests, compared with at
most 48.3\% and 27.1\%.

\paragraph{Latency--quality frontier.}
On KodCode (Figure~\ref{fig:frontier}), the default configuration of
\ours{} dominates every evaluated baseline configuration on both
backbones: none is both faster and more accurate. The advantage originates in parallel unit
decoding: because each iteration advances all units at once, \ours{}
completes the units of a KodCode program in 54 (Dream-Coder) and 61
(DiffuCoder) iterations on average, whereas the fixed-iteration
baselines denoise a canvas of about 1.5K positions for up to 512
iterations.

\subsection{Analysis}
\label{sec:exp-analysis}

\paragraph{Diffused versus left-to-right contracts.}
To isolate the benefit of diffusion-based planning, we let the same
weights write the contract left to right, as in Planned
Diffusion~\citep{israel2026planned}(Table~\ref{tab:schedules}). Diffused contracts
raise KodCode pass@1 from 22.1 to 25.8 on Dream-Coder and from 19.6 to
21.7 on DiffuCoder; they are also finer grained, with 6.7 instead of 5.3
entries on average, and all 240 of them parse. Diffusion thus yields
better decompositions, which supports using it for both planning and
implementation.

\paragraph{Binding and coupled attention.}
Table~\ref{tab:analysis} varies the two mechanisms that inform each
canvas's predictions (R3). Without binding, a canvas is tied to its
entry only by position, as in Planned Diffusion, and under isolated
attention 32.3\% of canvases begin with a sibling's definition. Binding
nearly eliminates misbinding and overruns and triples pass@1 under
isolated attention (4.2 to 13.3) at unchanged latency. Coupled attention
partly compensates for missing binding and adds 0.9 points on top of it.

\paragraph{Commitment threshold.}
Relative to single commitment, $\gamma_u=0.99$ reduces unit-decoding
iterations by 2.2--2.4$\times$ and end-to-end latency on held-out
KodCode by 1.65$\times$ and 1.82$\times$, with accuracy changes within
noise (25.8 to 24.2 and 21.7 to 23.3; sign test $p\geq0.52$). Lower
thresholds collapse, as R2 predicts: at $\gamma_u=0.7$, pass@1 falls
below 5\%, and the share of syntactically valid programs on Dream-Coder
drops from 96.7\% to 40.0\% (Table~\ref{tab:threshold}).
Appendix~\ref{app:results} reports further ablations.

\section{Conclusion}
\label{sec:conclusion}

We presented Self-Decomposing Diffusion for Code (\ours{}), a fully diffusion-based framework that integrates program decomposition and parallel implementation into a unified generative process. By explicitly modeling program structure and inter-unit dependencies through a generated contract, \ours{} enables coordinated parallel decoding while improving program correctness and substantially reducing latency. Our results across diverse benchmarks and diffusion backbones demonstrate that \emph{planning and parallelism can be complementary rather than competing objectives}. More broadly, this work suggests that realizing the potential of diffusion models for complex code generation requires not only faster token generation, but also explicit mechanisms for organizing and coordinating interdependent decisions. Future work could extend contract-guided diffusion to repository-scale generation and explore adaptive decomposition guided by execution feedback.

\section*{Limitations}
Our experiments use two open 7B diffusion code models. Diffusion models
still trail frontier autoregressive models, more capable diffusion
systems such as Gemini Diffusion~\citep{google2025geminidiffusion},
Mercury 2.5~\citep{ermon2026mercury25}, and Seed Diffusion
Preview~\citep{song2025seeddiffusion} do not release their weights, and
larger open diffusion models for code remain scarce; we expect stronger
backbones to benefit \ours{}, but we have not verified this. Contracts
capture nominal interfaces, namely the identifiers that units define and
require, but not types or behavioral specifications, and our supervision
relies on static analysis of Python programs, so other languages and
richer interfaces require corresponding analyzers. Finally, contract
synthesis adds a sequential stage before unit decoding. This overhead is
more than offset by parallel unit decoding, but it accounts for 4.5 of
the 7.6 seconds per KodCode problem with Dream-Coder, so faster contract
decoding would reduce latency further.

\section*{Ethical Considerations}
SDDC improves both accuracy and the efficiency of diffusion-based code generation, but it inherits the potential risks of automated code generation, including producing insecure or incorrect code and facilitating the development of harmful software. Generated programs should therefore be reviewed and tested before deployment, particularly in security-sensitive settings. Although faster decoding may reduce inference costs, training and deploying these models still consume computational resources and energy.

\bibliographystyle{unsrtnat}
\bibliography{custom}

\appendix
\section{Implementation Details}
\label{app:impl}

\subsection{Training Data}
\label{app:data}

\paragraph{KodCode programs.}
We start from the 487K problems of KodCode-V1~\citep{xu2025kodcode},
each a triple of question, reference solution, and unit tests. We keep
solutions that define at least three top-level functions or a class with
at least four methods and span at least 30 non-empty lines (29,047
problems), discard the competitive-programming subsets (22,418), discard
problems whose benchmark-similarity score provided by KodCode is at
least 0.75 (21,729), and re-execute every solution's tests in an
isolated sandbox, keeping solutions that pass (17,795). After
decomposition checks, the pool contains 17,499 programs. Problems in our
held-out and validation splits are excluded from training.

\paragraph{Unit annotation.}
Segmentation follows Section~\ref{sec:training}: class headers with their
constructors, methods, and top-level functions form segments, short
function-level programs may also be divided into statement segments, and
the segments concatenate exactly to the source. For a curated subset, an LLM
annotator read each program and its segments and, following a
written guideline, either rejected the program as unsuitable (for
example, interactive programs, programs dominated by placeholder code,
or programs that communicate through mutable global state) or partitioned
its segments into contiguous units. The guideline keeps every helper
that a sibling calls in its own unit, since its name must appear in the
contract, keeps public entry points separate, merges trivial unreferenced
helpers and symmetric families of near-identical methods, and isolates
trailing demonstration code. Every label is validated automatically,
requiring that the partition covers every segment exactly once, that
groups are contiguous, and that no group crosses a class boundary. On
252 programs labeled independently twice, the keep-or-reject decision
agrees in 94.8\% of cases and the exact partition in 77.1\%, and the
decisions on all 11 hidden reference items match their reference labels.
The annotation yields 2,770 programs. The remaining 13,203 pool programs
use one unit per segment, a rule that reproduces the annotated partition
exactly for 64.4\% of 1,200 annotated programs.

\paragraph{Example construction.}
The budget of unit $k$ is $b_k=\lceil(|h_k|+|\vx^{(k)}|+s_k)/\nu\rceil$,
where $|\cdot|$ counts tokens and $s_k$ is a small slack.
For every example, trailing demonstration or test code (assertions,
\texttt{\_\_main\_\_} guards, print-only statements, and test functions) is
removed, the units must reconstruct the program exactly, the budgets must
parse back exactly, and the example must fit into 2,048 tokens. The
annotated programs are upsampled threefold, giving 25,513 unit-view
examples; the 25,118 examples whose contracts fit into $P=224$ positions
also form contract-view examples. We hold out 10\% of the examples for
monitoring and train on the remaining 45,438.

\subsection{Training}
\label{app:training}

Table~\ref{tab:hparams} lists the hyperparameters, which are shared by
both backbones; Diffusion SFT uses the same settings and, without
contract-view examples, takes fewer steps. In the contract view, the
specification and the contract canvas attend bidirectionally. In the
unit view, the context $[\vc;\operatorname{ser}(\vz^\star)]$ is causal,
and canvas positions attend to the context and to all canvases (coupled)
or to their own canvas (isolated). Both backbones follow the shifted
prediction convention of Dream~\citep{ye2025dream}, in which the token
at position $j$ is predicted from the output at position $j-1$; we
therefore select the supervised outputs by the masking state of their
targets rather than of their inputs, which would leave a third of the
targets visible to the model. The next-token loss
$\mathcal{L}_{\textsc{ctx}}$ on contract tokens shares the $1/t$ weight
of its example. Diffusion SFT trains on the same 25,513 programs with
contract and canvas delimiters removed, applying the masked diffusion
objective to the full response with the same supervision indexing.

\begin{table}[t]
\caption{Training hyperparameters. The optimizer is paged 8-bit AdamW~\citep{loshchilov2019decoupled,dettmers2022bit}.}
\label{tab:hparams}
\centering
\small
\begin{tabular}{@{}ll@{}}
\toprule
Optimizer & paged 8-bit AdamW \\
Learning rate & $10^{-5}$ \\
Schedule & cosine, 3\% warm-up \\
Effective batch size & 32 \\
Epochs (steps) & 6 (8,520) \\
Sequence length & 2,048 tokens \\
Precision & bf16 \\
GPUs & 4$\times$ RTX PRO 6000 (96\,GB) \\
Training time & about 13 hours \\
\bottomrule
\end{tabular}
\end{table}

\subsection{Decoding}
\label{app:decoding}

\paragraph{\ours{}.}
On DiffuCoder, the admissible set $\mathcal{A}_j(\vy)$ also contains
$\pad$ directly after a complete entry, and a $\pad$ committed there ends
the contract. Unit decoding samples candidates with temperature 0.2 and
top-$p$ 0.95. The keys and values of the context are computed once, and
those of completed canvases are reused; under coupled attention this
reuse is an approximation, because they were computed while other
canvases were still changing.

\paragraph{Baselines.}
Vanilla and Diffusion SFT denoise a canvas that fills the length budget
of each benchmark, including the prompt: 2,048 tokens for KodCode and
ClassEval, 4,096 for ClassEval-Pro, 1,280 for HumanEval Pro, and 1,024
for MBPP Pro. Fixed-iteration decoding uses the entropy-ordered sampler
of Dream~\citep{ye2025dream} with temperature 0.2 and top-$p$ 0.95.
Threshold decoding commits every position whose confidence is at least
0.8, and at least one per iteration, for up to 512 iterations. We
evaluate Vanilla with 128, 256, and 512 iterations and Diffusion SFT
with 64, 128, and 256 iterations on KodCode, both with 256 iterations on
the other benchmarks and Vanilla additionally with 512, and both with
threshold decoding everywhere. The most accurate of these configurations
is 512 iterations for Vanilla on every benchmark and threshold decoding
for Diffusion SFT, except on KodCode and HumanEval Pro with DiffuCoder,
where 256 iterations are best; Tables~\ref{tab:program}
and~\ref{tab:selfinvoking} report these configurations. On the
self-invoking benchmarks, we further evaluate Diffusion SFT with
thresholds 0.9 and 0.95 and with 512 iterations
(Table~\ref{tab:sft-sweep}). For baselines, the program is the fenced
code block of the answer, or all blocks that define code and parse if
there are several, or the whole answer if there is none.

\paragraph{Hyperparameter selection.}
The commitment threshold $\gamma_u$ and the DiffuCoder contract variant
were selected on the 210-problem validation split. For Dream-Coder, the
canvas scale $\lambda$ and the contract schedule were chosen on KodCode
during development; the configuration that is best on the validation
split alone ($\lambda=16$) attains the same held-out pass@1 within one
problem.

\subsection{Evaluation}
\label{app:eval}

\paragraph{Benchmarks.}
Every method receives the benchmark prompt as the user turn of its chat
template. The KodCode held-out split contains 240 problems with 120
multi-function modules and 120 class-based programs, 48 of which define
several classes, and 80 problems at each difficulty level; the 210
validation problems are drawn in the same way. KodCode programs are run
against their pytest suites with a timeout of 300 seconds. On ClassEval,
the class pass rate requires all unit tests of a class to pass, and the
test pass rate is the fraction of all unit tests passed. On
ClassEval-Pro, whose reference solutions have a median length of 1,001
tokens, we use the official holistic prompt and evaluator: its class
success requires all test classes of a class to pass, and its
function-level success, our test pass rate, is the fraction of test
classes whose tests all pass. Because some official tests patch
built-in functions, we run the evaluator separately for each problem and
sample, so that a failing program cannot affect later ones. On HumanEval
Pro and MBPP Pro, we use the official sanitizer and harness and report
the official score, the better of raw and sanitized outputs; every
method receives the official instruction as a chat turn rather than a
pre-filled function signature, because \ours{} begins its response with
a contract.

\paragraph{pass@5.}
We draw five samples per problem at temperature 0.4 and top-$p$ 0.95. All
stochastic stages sample, including contract synthesis and threshold
decoding, whose commitment decisions use the untempered probability of
the sampled token so that the thresholds stay calibrated. A problem
counts as solved if any sample passes; on ClassEval, a unit test counts
as passed if any sample passes it, and on ClassEval-Pro we use the
official pass@$k$ estimator. Sampling at temperature 0.4 lowers the
per-sample accuracy of several models, most strongly of the base models,
which then often answer in prose without code; pass@5 can therefore fall
below pass@1 at temperature 0.2.

\paragraph{Latency.}
Latency is the wall-clock time per problem with batch size one,
including contract synthesis and context encoding, measured on a single
NVIDIA RTX PRO 6000 GPU of the same machine for all methods. The cost of
fixed-iteration decoding varies by less than 5\% across problems, so we
time it on subsets of 20--60 problems. For pass@5, latency is the time
to draw all five samples.

\section{Proofs}
\label{app:proofs}

\paragraph{Proposition~\ref{prop:step}.}
Write $p_\star$ for $p_\star(\cdot\mid\eta)$, $H_\star$ for entropy
under $p_\star$, and $p_{\theta,i}$ for $p_{\theta,i}(\cdot\mid\eta)$.
Adding and subtracting
$\sum_{i\in Q}\log p_\star(X^i)$ inside the expectation gives
\begin{equation*}
\begin{aligned}
&\KL\Bigl(p_\star(X^Q)\,\Big\|\prod_{i\in Q}p_{\theta,i}(X^i)\Bigr)\\
&\quad=\E\Bigl[\log\frac{p_\star(X^Q)}{\prod_ip_\star(X^i)}\Bigr]
+\sum_{i\in Q}\E\Bigl[\log\frac{p_\star(X^i)}{p_{\theta,i}(X^i)}\Bigr]\\
&\quad=\operatorname{TC}_\star(X^Q)
+\sum_{i\in Q}\KL\bigl(p_\star(X^i)\,\|\,p_{\theta,i}(X^i)\bigr),
\end{aligned}
\end{equation*}
with expectations under $p_\star(X^Q)$. Adding and subtracting
$\sum_kH_\star(X^{Q_k})$ in the definition of total correlation yields
\begin{equation*}
\operatorname{TC}_\star(X^Q)
=\sum_k\operatorname{TC}_\star(X^{Q_k})
+\operatorname{TC}_\star\bigl((X^{Q_k})_k\bigr),
\end{equation*}
because $\sum_{i\in Q}H_\star(X^i)-\sum_kH_\star(X^{Q_k})$ is the sum of the
within-unit total correlations and
$\sum_kH_\star(X^{Q_k})-H_\star(X^Q)$ is the total correlation among the
units. Substituting this identity proves Eq.~\eqref{eq:error-decomposition}.
\hfill$\square$

\paragraph{Confidence bound.}
Fix a unit $k$ and any position $j\in Q_k$. Because
$H_\star(X^{Q_k})\geq H_\star(X^j)$, the within-unit term satisfies
\begin{equation*}
\begin{aligned}
\operatorname{TC}_\star(X^{Q_k})
&=\sum_{i\in Q_k}H_\star(X^i)-H_\star(X^{Q_k})\\
&\leq\sum_{i\in Q_k\setminus\{j\}}H_\star(X^i).
\end{aligned}
\end{equation*}
If the most probable token at position $i$ has probability
$p_i\geq\gamma\geq\frac12$, grouping the remaining tokens gives
$H_\star(X^i)\leq H_{\mathrm b}(p_i)+(1-p_i)\log(|\gV|-1)$, and the
right-hand side decreases in $p_i$ on $[\frac12,1]$. Hence each position
of $Q_k$ other than $j$ adds at most
$H_{\mathrm b}(\gamma)+(1-\gamma)\log(|\gV|-1)$ to the within-unit term,
a bound that vanishes as $\gamma\to1$. No condition is required at $j$,
which corresponds to the single sampled commitment of each canvas in
Algorithm~\ref{alg:ours}. \hfill$\square$

\section{Additional Results}
\label{app:results}

\paragraph{Decoding configurations of Diffusion SFT.}
Table~\ref{tab:sft-sweep} evaluates Diffusion SFT on the self-invoking
benchmarks with five decoding configurations. Below 20 seconds per
problem, its best configuration remains 12.1--20.8 points below \ours{};
512 iterations, its most accurate configuration, costs 35--40 seconds
per problem and surpasses \ours{} only on MBPP Pro with Dream-Coder, by
1.6 points at about eight times the latency.

\begin{table}[t]
\caption{Diffusion SFT under different decoding configurations on HumanEval Pro and MBPP Pro: pass@1 (\%), with the mean latency per problem in seconds in the \textcolor{blue!70!black}{blue} row below. Vanilla (512 iterations) and \ours{} are shown for reference; the most accurate entry of each column is \textbf{bolded}.}
\label{tab:sft-sweep}
\centering
\renewcommand{\lat}[1]{\textcolor{blue!70!black}{#1}}%
\renewcommand{\arraystretch}{1.05}
\setlength{\tabcolsep}{4pt}
\providecommand{\narrowtablewidth}{\columnwidth}%
\resizebox{\narrowtablewidth}{!}{%
\begin{tabular}{@{}l llll@{}}
\toprule
& \multicolumn{2}{c}{\textbf{HumanEval Pro}} & \multicolumn{2}{c}{\textbf{MBPP Pro}} \\
\cmidrule(lr){2-3} \cmidrule(lr){4-5}
\textbf{Decoding} & Dream-Coder & DiffuCoder & Dream-Coder & DiffuCoder \\
\midrule
Threshold 0.8   & 26.8 & 14.0 & 36.0 & 31.5 \\
                & \lat{4.2} & \lat{5.5} & \lat{2.3} & \lat{2.8} \\
Threshold 0.9   & 31.1 & 15.2 & 40.2 & 33.3 \\
                & \lat{6.6} & \lat{9.2} & \lat{3.7} & \lat{4.6} \\
Threshold 0.95  & 30.5 & 14.6 & 40.5 & 33.9 \\
                & \lat{9.0} & \lat{12.7} & \lat{5.4} & \lat{6.5} \\
256 iterations  & 15.2 & 20.1 & 24.6 & 28.0 \\
                & \lat{19.8} & \lat{19.7} & \lat{17.3} & \lat{17.3} \\
512 iterations  & 33.5 & 37.2 & \textbf{54.2} & 49.5 \\
                & \lat{39.5} & \lat{39.5} & \lat{34.7} & \lat{34.7} \\
\midrule
\textit{Vanilla} & 40.2 & 31.1 & 51.9 & 48.1 \\
                 & \lat{39.4} & \lat{39.5} & \lat{34.7} & \lat{34.7} \\
\textit{\ours{}} & \textbf{44.5} & \textbf{40.9} & 52.6 & \textbf{49.7} \\
                 & \lat{5.1} & \lat{3.2} & \lat{4.4} & \lat{3.2} \\
\bottomrule
\end{tabular}}
\end{table}

\paragraph{Contract schedules.}
Table~\ref{tab:schedules} compares contract schedules on KodCode with a
single commitment per canvas. Besides the default schedule
($S_z=112$, padding only as a suffix), \emph{end-after-entry} also admits
$\pad$ directly after a complete entry, and \emph{pad-scan} ($S_z=56$)
extends the padding suffix in each iteration over all masked positions
whose padding probability exceeds $\gamma_z$. Faster schedules shorten
contracts and trade accuracy for latency; the default schedule is the
most accurate on Dream-Coder, and all three diffused schedules
outperform left-to-right synthesis on DiffuCoder. With $\gamma_u=0.99$,
the configurations of Table~\ref{tab:program} reach 24.2 and 23.3 pass@1
at 7.55 and 5.26 seconds, of which contract synthesis takes 4.55 and
2.19 seconds and unit decoding, which needs 54 and 61 iterations on
average, 3.00 and 3.07 seconds.

\begin{table}[t]
\caption{Contract schedules on KodCode (held-out) with single commitment per canvas: pass@1 (\%) and latency (s).}
\label{tab:schedules}
\centering
\small
\setlength{\tabcolsep}{4pt}
\begin{tabular}{@{}l cc cc@{}}
\toprule
& \multicolumn{2}{c}{Dream-Coder} & \multicolumn{2}{c}{DiffuCoder} \\
\cmidrule(lr){2-3} \cmidrule(lr){4-5}
Contract & pass@1 & Lat. & pass@1 & Lat. \\
\midrule
Default & \textbf{25.8} & 12.46 & 22.1 & 12.37 \\
End-after-entry & 22.9 & 9.81 & 21.7 & 9.55 \\
Pad-scan & 21.7 & 8.56 & \textbf{22.5} & 8.90 \\
Left-to-right & 22.1 & 9.07 & 19.6 & 8.92 \\
\bottomrule
\end{tabular}
\end{table}

\paragraph{Commitment threshold.}
Table~\ref{tab:threshold} reports validation pass@1 and the mean number
of unit-decoding iterations for each commitment threshold $\gamma_u$.

\begin{table}[t]
\caption{Commitment threshold $\gamma_u$ on the KodCode validation split: pass@1 (\%) and mean unit-decoding iterations (Iter.); ``single'' commits one token per canvas per iteration.}
\label{tab:threshold}
\centering
\small
\setlength{\tabcolsep}{4pt}
\begin{tabular}{@{}l cc cc@{}}
\toprule
& \multicolumn{2}{c}{Dream-Coder 7B} & \multicolumn{2}{c}{DiffuCoder 7B} \\
\cmidrule(lr){2-3} \cmidrule(lr){4-5}
$\gamma_u$ & pass@1 & Iter. & pass@1 & Iter. \\
\midrule
single & 22.4 & 137 & 20.0 & 143 \\
0.99   & \textbf{24.8} & 56 & \textbf{20.5} & 64 \\
0.95   & 19.5 & 33 & 18.6 & 39 \\
0.90   & 17.1 & 25 & 15.2 & 28 \\
0.80   & 10.5 & 17 & 9.5  & 20 \\
0.70   & 4.8  & 14 & 3.8  & 16 \\
\bottomrule
\end{tabular}
\end{table}

\paragraph{Binding components.}
In the controlled setting of Table~\ref{tab:analysis} with isolated
attention, validation pass@1 is 5.2 without binding, 5.7 with
length-scaled padding alone, 9.5 with headers alone, and 12.9 with both;
configurations without length-scaled padding draw every surplus from
$\mathcal{U}\{0,\ldots,10\}$, and coupled attention with both components
reaches 14.8. Length-scaled padding halves the overruns
into the next unit, from 9.2\% to 4.7\% without headers and from 4.4\%
to 1.4\% with them, so the two components are complementary.

\paragraph{Canvas scale and pad-fill.}
In the same controlled setting, $\lambda=10$, 14, and 18 yield 14.2,
21.7, and 19.2 held-out pass@1 at 6.54, 10.86, and 16.06 seconds:
small canvases truncate units, whereas large ones slow decoding without
improving accuracy. Pad-fill exploits the contiguous padding suffix
induced by length-scaled padding; for a model trained on the full corpus with
left-to-right contracts and $\lambda=14$, it reduces latency from 10.42
to 8.55 seconds while pass@1 changes from 21.2 to 22.1.

\end{document}